\documentclass[10pt,letterpaper]{article}

\usepackage{cogsci}

\cogscifinalcopy 

\usepackage{pslatex}
\usepackage{apacite}
\usepackage{float} 

\usepackage{graphicx}
\usepackage{multirow}
\usepackage{booktabs}
\usepackage{amsfonts}       
\usepackage{amsmath}        
\usepackage{algorithm}      
\usepackage{algpseudocode}  
\usepackage{listings}       
\usepackage{enumitem}       
\usepackage{xcolor}         
\usepackage{subcaption}     
\usepackage{babel}
\title{Grounding Language about Belief in a Bayesian Theory-of-Mind}

\author{\\\\{\large \bf Lance Ying$^{*12}$, Tan Zhi-Xuan$^{*1}$, Lionel Wong$^{1}$, Vikash Mansinghka$^{1}$, Joshua B. Tenenbaum$^{1}$} \\
  $^{1}$Massachusetts Institute of Technology, Cambridge, MA, USA \\
  $^{2}$Harvard University, Cambridge, MA, USA
  }

\begin{document}
\maketitle

\begin{abstract}
Despite the fact that beliefs are mental states that cannot be directly observed, humans talk about each others' beliefs on a regular basis, often using rich compositional language to describe what others think and know. What explains this capacity to interpret the hidden epistemic content of other minds? In this paper, we take a step towards an answer by grounding the semantics of belief statements in a Bayesian theory-of-mind: By modeling how humans jointly infer \emph{coherent} sets of goals, beliefs, and plans that explain an agent's actions, then evaluating statements about the agent's beliefs against these inferences via epistemic logic, our framework provides a functional role semantics for belief, explaining the gradedness and compositionality of human belief attributions, as well as their intimate connection with goals and plans. We evaluate this framework by studying how humans attribute goals and evaluate belief sentences while watching an agent solve a doors-and-keys gridworld puzzle that requires instrumental reasoning about hidden objects. In contrast to pure logical deduction, non-mentalizing baselines, and mentalizing that ignores the role of instrumental plans, our model provides a much better fit to human goal and belief attributions, demonstrating the importance of theory-of-mind for modeling how humans understand language about beliefs.

\end{abstract}
\let\thefootnote\relax\footnotetext{$^*$Equal Contribution}

\section{Introduction}

Language about belief is pervasive in social life. Whether one is telling a friend about a mutual friend's aspirations (\emph{"She seems to think she'll get the role she's applying for..."}), speculating about the insider knowledge behind an investment decision (\emph{"Meta probably thinks this technology is going to be a big deal..."}), or listening to news commentators explain the actions of governments (\emph{"The UK believes that this policy will improve..."}), language provides an incredibly rich medium for communicating about our mental models of the world. Remarkably, we often talk about belief despite not knowing what other people truly think, instead inferring beliefs from what they say or do. Belief, after all, is a guide to rational action, and there are many things a person would or would not do if they believed some fact to be true.

How do people understand language about belief, in light of this intimate connection? Put another way, what do people take as the \emph{meaning} of belief statements, given the tight links between language, thought, and action? While philosophers, logicians, and linguists have long studied the semantics of belief \shortcite{gochet2006epistemic,chisholm1955sentences,loar1981mind}, many of these inquiries have focused on the relationship between a belief sentence and the proposition it embeds \shortcite{partee1973semantics,stalnaker1987semantics} or how belief sentences can be nested and updated \shortcite{bolander2017gentle}, not how people use and interpret such sentences in everyday social contexts. On the other hand, computational cognitive scientists have developed numerous models that explain how people attribute beliefs and goals to other agents \shortcite{baker2009action,jara2019naive,houlihan2023emotion,poppel2019satisficing,zhang2023goal}. Building upon the work of \shortciteA{baker2017rational}, these models formulate mental state attributions as the product of a Bayesian theory-of-mind (BToM): Since beliefs and goals play a \emph{functional} role in guiding plans and actions \cite{dennett1989intentional}, it is possible to \emph{interpret} an agent's actions as driven by particular goals and beliefs through Bayesian inference. However, prior work in this paradigm has not studied the rich compositional nature of belief sentences, or how humans might interpret them.

\begin{figure*}[t]
    \centering
    \includegraphics[width=0.7\textwidth]{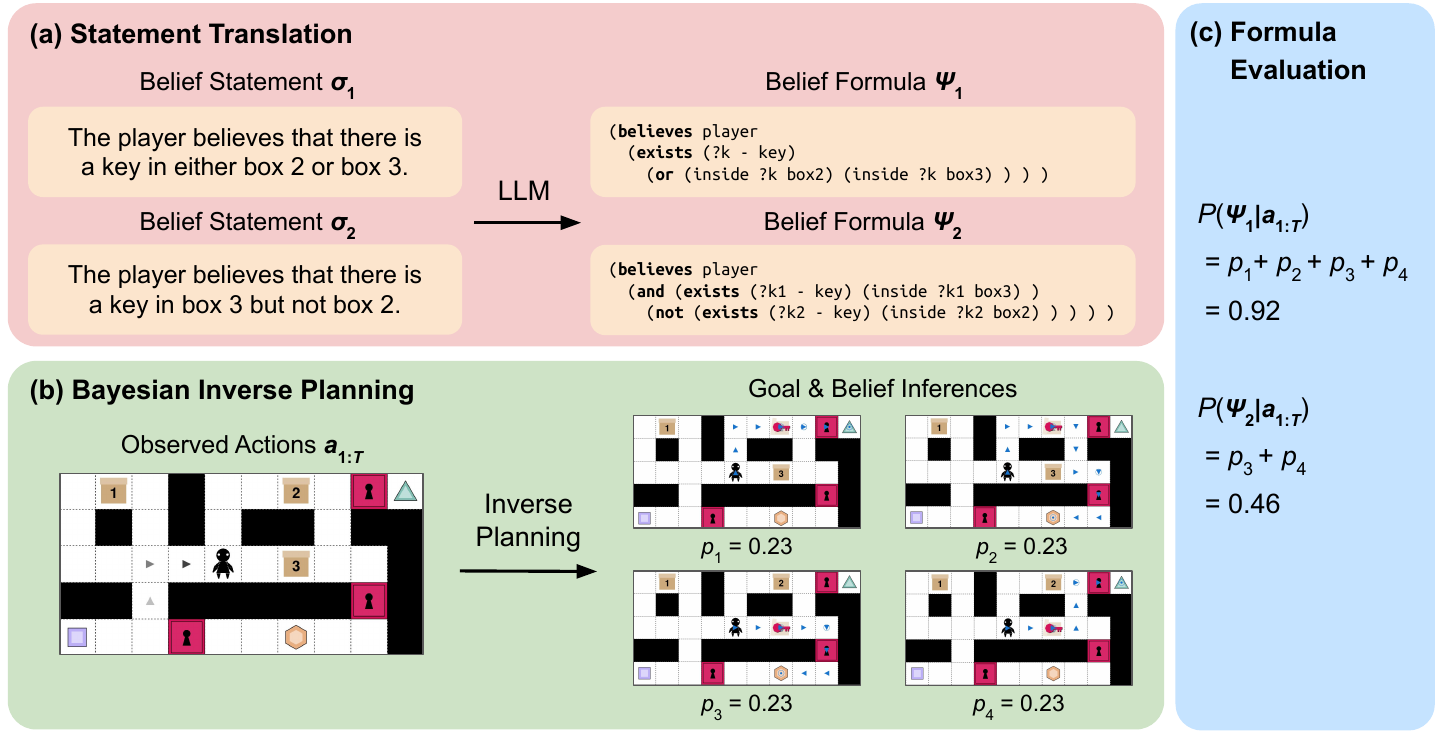}
    \caption{Overview of our proposed model. \textbf{(a)} We translate belief statements $\sigma_i$ from English into first-order epistemic logic (using a large language model). \textbf{(b)} Given a series of observed actions $a_{1:T}$, we simulate possible mental states (combinations of the agent's goals, beliefs, and plans), assigning a posterior probability $p_i$ to each hypothesis via inverse planning. \textbf{(c)} Each logical formula $\psi_i$ can then be evaluated against the inferred mental states to produce a probability rating $P(\psi_i|a_{1:T})$.}
    \label{fig:illustration}
\end{figure*}
In this paper, we ground \emph{natural language statements} of belief in the functional role semantics \cite{harman1982conceptual} afforded by a Bayesian theory-of-mind. In our approach, belief statements are not just claims about which propositions are thought to be true by an agent --- which we capture with epistemic logic \cite{gochet2006epistemic} --- but also imply \emph{rationality constraints} on how the agent is likely to plan and act given their goals \cite{loar1981mind} --- as captured by our Bayesian agent model. As such, our framework explains how humans can assess the likeliness of a belief statement based on how well it coheres with an agent's actions, along with the goals and intentions those actions imply. To implement this framework, we combine the strengths of machine learning methods with the coherence and precision of Bayesian and logical reasoning \cite{wong2023word, ying2023neuro}, using a large language model (LLM) as a tool to automatically translate natural language statements into logical form \shortcite{shin2021constrained}, then evaluating these statements with respect to the inferences produced by a probabilistic programming architecture for Bayesian inverse planning \shortcite{zhi2020online}.

To evaluate this framework, we design an experiment where participants are shown animations of a player navigating a gridworld puzzle, which requires the player to collect one of four valuable gems (see Figure \ref{fig:illustration}). These gems are sometimes locked behind doors, so the player may have to collect keys located inside opaque boxes in order to reach their desired gem. For simplicity, we assume that the player \emph{knows} where the keys are located. However, participants observing the scenario do not. As such, they have \emph{infer} the player's goals and beliefs from the actions they observe, then provide their inferences as ratings. By comparing our model's outputs to human ratings, we evaluate how well it explains people's interpretations of language about belief.


\section{Computational Model}

To explain how human observers interpret and evaluate language about belief, our model makes use of: (i) epistemic logic as a formal compositional representation of belief statements; (ii) a Bayesian generative model that encodes the functional role that belief plays 
\emph{vis a vis} an agent's goals, plans, and actions (i.e., a Bayesian ToM). These two components allow us to (iii) infer likely goals and beliefs from an agent's actions, then (iv) assess the likeliness of a belief statement with respect to those inferences.

\subsection{Representing Belief Statements}

As an expressive representation of belief statements and the environments they describe, we add epistemic modalities to the Planning Domain Definition Language (PDDL), a first-order language for model-based planning and reasoning \cite{mcdermott1998pddl}. In a PDDL domain, a set of predicates $\mathcal{P}$ are used to describe objects $\mathcal{O}$, each of which has a type $\tau \in \mathcal{T}$. For example, to represent the fact that \texttt{key1} is red in color, we write \texttt{(iscolor key1 red)}, with predicate $\texttt{iscolor} \in \mathcal{P}$ and objects $\texttt{key1}, \texttt{red} \in \mathcal{O}$, with types $\texttt{key}, \texttt{color} \in \mathcal{T}$ respectively. Each environment state $s$ is essentially a set of such predicate terms, stating which relations are true or false between objects.

Since PDDL is first-order, we can evaluate the truth value of a sentence $\phi$ in a state $s$, where $\phi$ is compositionally defined in terms of logical operations and quantifiers, and predicates can take objects $o \in \mathcal{O}$ or variables \texttt{?v} as arguments $x_i$:
\begin{align*}
\phi ::= &\texttt{($P$ $x_1$ $...$ $x_n$)} | \texttt{(not $\phi$)} | \texttt{(and $\phi_1$ $\phi_2$)} | \texttt{(or $\phi_1$ $\phi_2$)} | \\
&\texttt{(exists (?v - $\tau$) $\phi$)} | \texttt{(forall (?v - $\tau$) $\phi$)}
\end{align*}
This expressivity allows us to determine the truth value of not just individual relations, but also general queries about whether some property holds for some class of objects.

However, PDDL alone cannot express claims about what an agent \emph{believes}. To do this, we follow work in epistemic planning \shortcite{bolander2017gentle,muise2022efficient}, introducing a \texttt{believes} operator: Given some regular PDDL sentence $\phi$ and agent $x$, \texttt{(believes $x$ $\phi$)} means that agent $x$ believes $\phi$. As such, a statement like \emph{``The player believes that there is a red key in box 1''} can be represented as  \texttt{(believes player (exists (?k - key) (and (iscolor ?k red) (inside ?k box1)))}. This extended language corresponds to a restricted fragment of epistemic first-order logic \cite{gochet2006epistemic}.
 
\subsection{Modeling the Functional Role of Belief}

Epistemic logic allows us to express compositional belief statements. But what do such beliefs imply about an agent's likely behavior? We formally model this functional connection with a probabilistic generative model:
\begin{alignat}{2}
\textit{Goal Prior:}& \quad g \sim P(g) \label{eq:goal-prior} \\
\textit{State Prior:}& \quad s_0 \sim P(s_0) \label{eq:state-prior} \\
\textit{Belief Update:}& \quad b_t \sim P(b_t | s_t) \label{eq:belief-update} \\
\textit{Action Selection:}& \quad a_t \sim P(a_t | b_{t-1}, g) \label{eq:action-selection} \\
\textit{State Transition:} & \quad s_t \sim P(s_t | s_{t-1}, a_t) \label{eq:state-transition}
\end{alignat}
In this model, we assume that agents are usefully described as having beliefs $b_t$ and goals $g$ insofar as they take actions $a_t$ towards the goal that are (approximately) rational given their beliefs. As such, our model defines a \emph{functional role} for belief. Notably in our setting, we focus on the simpler case where the agent has both \emph{complete knowledge} and \emph{veridical perception}, such that their belief at each step $t$ is always equivalent to the environment state $s_t$. Nonetheless, an observer will still have uncertainty over what the agent believes, since they have uncertainty over the initial state $s_0$.

What does it mean for an agent's actions to be rational with respect to their beliefs (and goals)? Following the principle of rational action \shortcite{gergely2003teleological}, this means that actions should lead efficiently to a goal in the world that the agent believes it is in. In a multi-step task like ours (see Figure \ref{fig:illustration}), an agent should thus \emph{plan} to reach any instrumental subgoals necessary for their final goal (e.g. keys in boxes), then act roughly according to that plan. We capture this probabilistically by adopting a Boltzmann-rational model of action selection, given the agent's goal $g$ and their belief $b_{t-1}$:
\begin{equation}
    P(a_t | b_{t-1}, g) = \frac{\exp\left(-\beta \hat Q_g(b_{t-1}, a)\right)}{\sum_{a'} \exp\left(-\beta \hat Q_g(b_{t-1}, a')\right)}
\end{equation}
Here $\hat Q_g(b_{t-1}, a)$ represents the estimated cost of the optimal plan to achieve goal $g$ after taking action $a$ starting at the believed state $b_{t-1}$. $\beta$ is a parameter controlling action optimality (higher is more optimal). To compute $\hat Q_g(b_{t-1}, a)$ efficiently, we leverage recent advances in sequential inverse planning \shortcite{zhi2020online,zhixuan2024pragmatic}, using real-time heuristic search \shortcite{koenig2006real,hernandez2011tree} to rapidly estimate the cost to the goal $g$.

\subsection{Joint Inference of Goals and Beliefs}

Similar to \citeA{baker2017rational} and others \cite{farrell2020narrative}, we model observers as performing joint inference over the agent's goal $g$ and belief history $b_{0:T}$ given a series of observed actions $a_{1:T}$, The corresponding posterior $P(g, b_{0:T} | a_{1:T})$ factorizes as follows:
\begin{multline}
    P(g, b_{0:T} | a_{1:T}) \propto \textstyle\sum_{s_{0:T}} P(g) P(s_0) P(b_0 | s_0) \\ \textstyle\prod_{t=1}^T P(a_t | b_{t-1}, g) P(s_t | s_{t-1}, a_t) P(b_t | s_t)
\end{multline}
Since we assume the agent has perfect knowledge of the environment ($b_t \equiv s_t$), this reduces to inferring $g$ and $s_{0:T}$:
\begin{multline}
    P(g, s_{0:T} | a_{1:T}) \propto \\ P(g) P(s_0) \textstyle\prod_{t=1}^T P(a_t | s_{t-1}, g) P(s_t | s_{t-1}, a_t)
\end{multline}

To compute this distribution, we initialize a set of weighted samples $\{(g^i, s_0^i, w^i)\}_{i=1}^{N}$ by enumerating over all possible combinations of goals and initial states\footnote{In settings where there are too many possibilities to enumerate over, we can sample a representative set of goals and states instead.}, where $N$ is the number of combinations, and $w^i$ is initialized to $P(g^i) P(s_0^i)$. Then we sequentially update each sample $i$, multiplying $w^i$ by the likelihood $P(a_t|s_{t-1}^i, g^i)$ of observing action $a_t$, and simulating the next state $s_t$ that results from $a_t$:
\begin{equation*}
    w^i \gets w^i \cdot P(a_t| s_{t-1}^i, g^i) \qquad
    s_t \sim P(s_t^i | s_{t-1}^i, a_t)
\end{equation*}
Each weight $w^i$ represents the \emph{unnormalized} probability of the pair $(g^i, s_{0:t}^i)$ at step $t$. Normalizing the weights gives us the probability $P(g^i, s_{0:t}^i | a_{1:t}) = w^i / \textstyle\sum_{j=1}^N w_j$

In stochastic environments, this procedure is a sequential Monte Carlo algorithm \shortcite{del2006sequential}. However, since our environment is deterministic, this algorithm reduces to \emph{exact} Bayesian filtering, which we implement efficiently as a variant of Sequential Inverse Plan Search \cite{zhi2020online} using the Gen probabilistic programming system \shortcite{cusumano2019gen}.

\subsection{Quantitatively Interpreting Belief Statements}

Having computed the posterior $P(g, b_{0:T} | a_{1:T})$, we model the interpretation of belief statements in two steps: First, we \emph{translate} the natural language statement $\sigma$ into a a formula $\psi$ in epistemic logic. Second, we \emph{evaluate} the expected truth value of the translated sentence $\psi$ with respect to inferred distribution over belief states $b_T$, thereby grounding the meaning of the sentence in our BToM model.

To perform the translation itself, it is possible to use a domain-specific grammar 
that maps English words to predicates and operators in our epistemic extension of PDDL. However, this can typically only handle a restricted fragment of natural language, while requiring significant engineering effort. Hence, we opt to use large language models (LLMs) as general purpose semantic parsers \cite{wong2023word}, which can translate natual language statements $\sigma$ to symbolic forms $\psi$ after being provided with only a few example translations \cite{shin2021constrained}. While not the focus of our present study, we expect this approach will enable us to approach human-level flexibility when interpreting language about belief.

Next, to evaluate the observer's credence in $\psi := \texttt{(believes $x$ $\phi$)}$, we extract the sentence $\phi$ attributed to the agent $x$. We then compute the expected value of $\phi$ being true in the belief state $b_T$: \begin{align*}
    P(\psi | a_{1:T}) &= \mathbb{E}_{b_T \sim P(b_T | a_{1:T})}[\phi \text{ is true in } b_T]  \\
    &= \textstyle\sum_{i=1}^N w_i \cdot [\phi \text{ is true in } b_T^i] / \textstyle\sum_{j=1}^N w_j
\end{align*}

One notable aspect of this formulation is that the value of $P(\psi | a_{1:T})$
depends on the prior over belief states $P(b_0) = P(s_0)$. However, it is not obvious what priors $P(b_0)$ human observers bring to bear. As such, we experiment with two possibilities: $U_{S_0}(s_0)$ is uniform over the set of possible initial states $s_0 \in S_0$, and $U_{\psi}(s_0)$ induces a uniform prior $P(\psi) = 0.5$ over the \emph{statement} being true. In the second case, $P(\psi | a_{1:T})$ becomes a kind of (normalized) likelihood $\bar L(\psi | a_{1:T}) = \frac{P(a_{1:T} | \psi)}{P(a_{1:T} | \psi) + P(a_{1:T} | \neg\psi)}$, which rates how much more evidence there is for the statement $\psi$ as opposed to its negation $\neg\psi$. In the absence of evidence, $\bar L(\psi | a_{1:T}) = 0.5$.


\section{Experiments}


To evaluate our model of how humans interpret belief statements, we designed an experiment where participants had to infer the goals and beliefs of an agent solving a gridworld puzzle called Doors, Keys, \& Gems \cite{zhi2020online}. In these puzzles, an agent has to pick up keys and unlock doors to reach a valuable gem. Doors can only be opened by keys of the same color and each key can be used once. To introduce partial observability, we also added boxes, each of which might be empty or contain exactly one key. 

\begin{figure*}[p]
    \centering

    \includegraphics[width=0.3\textwidth]{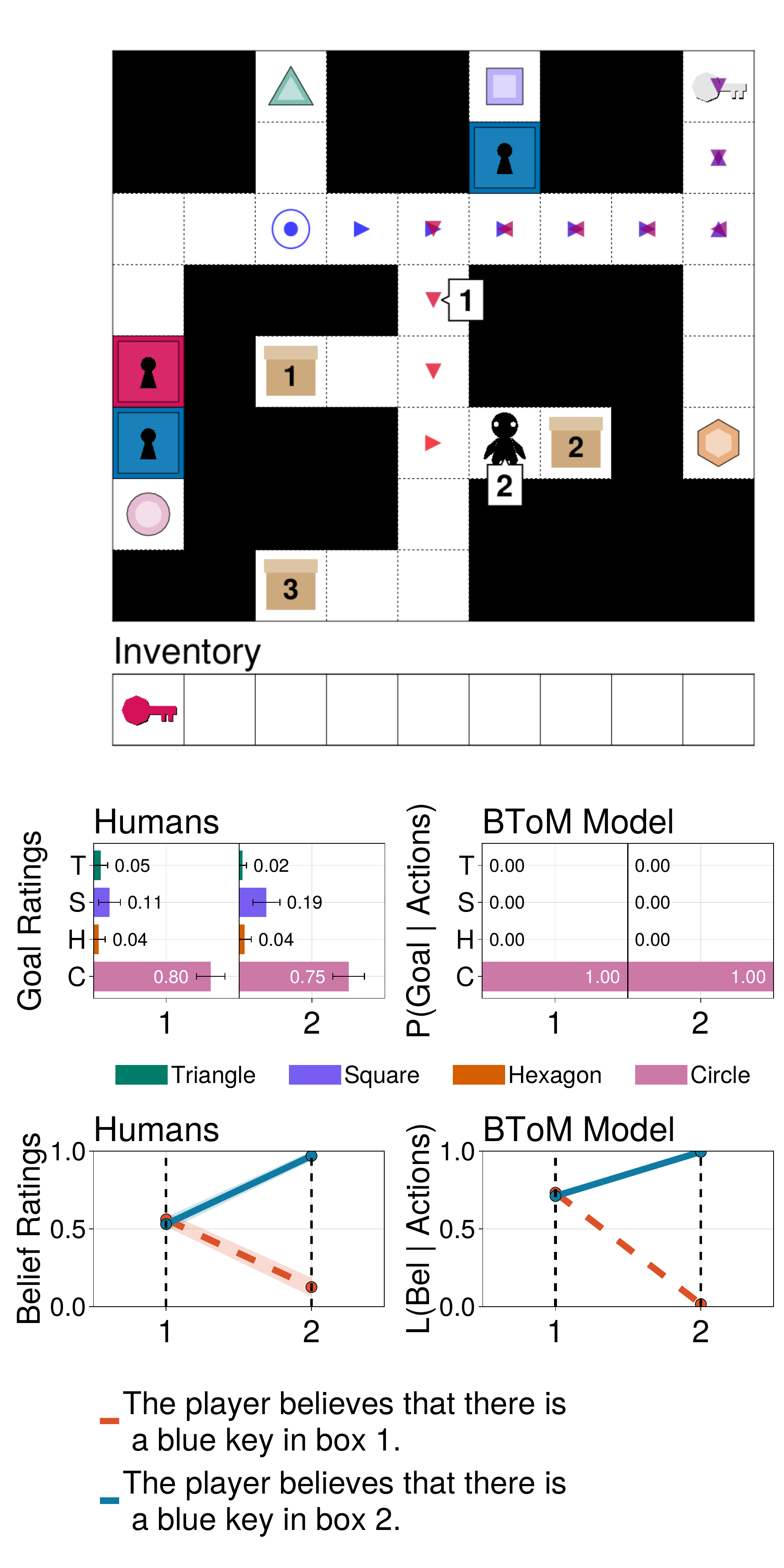}
    \includegraphics[width=0.3\textwidth]{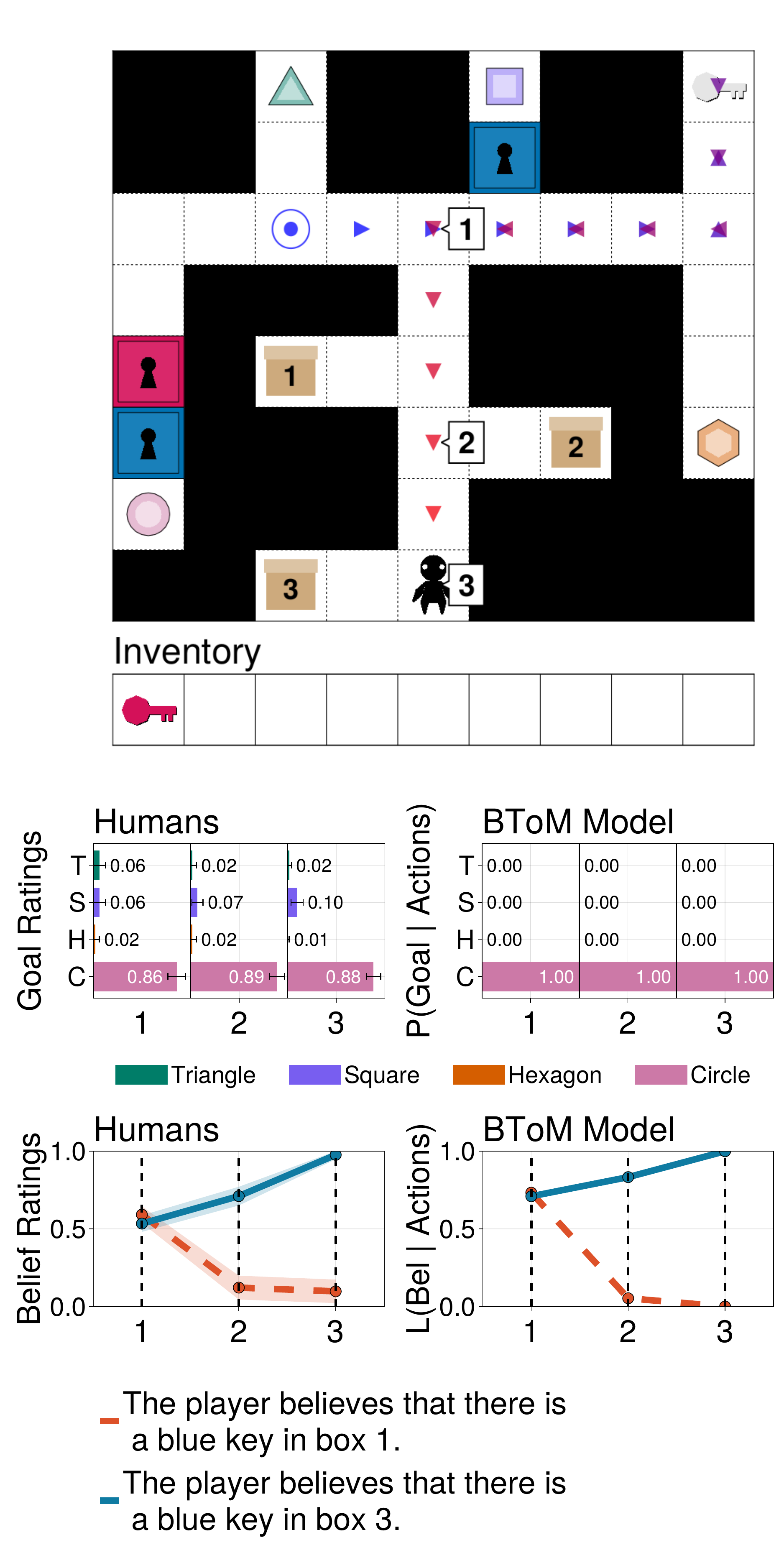}
    \includegraphics[width=0.3\textwidth]{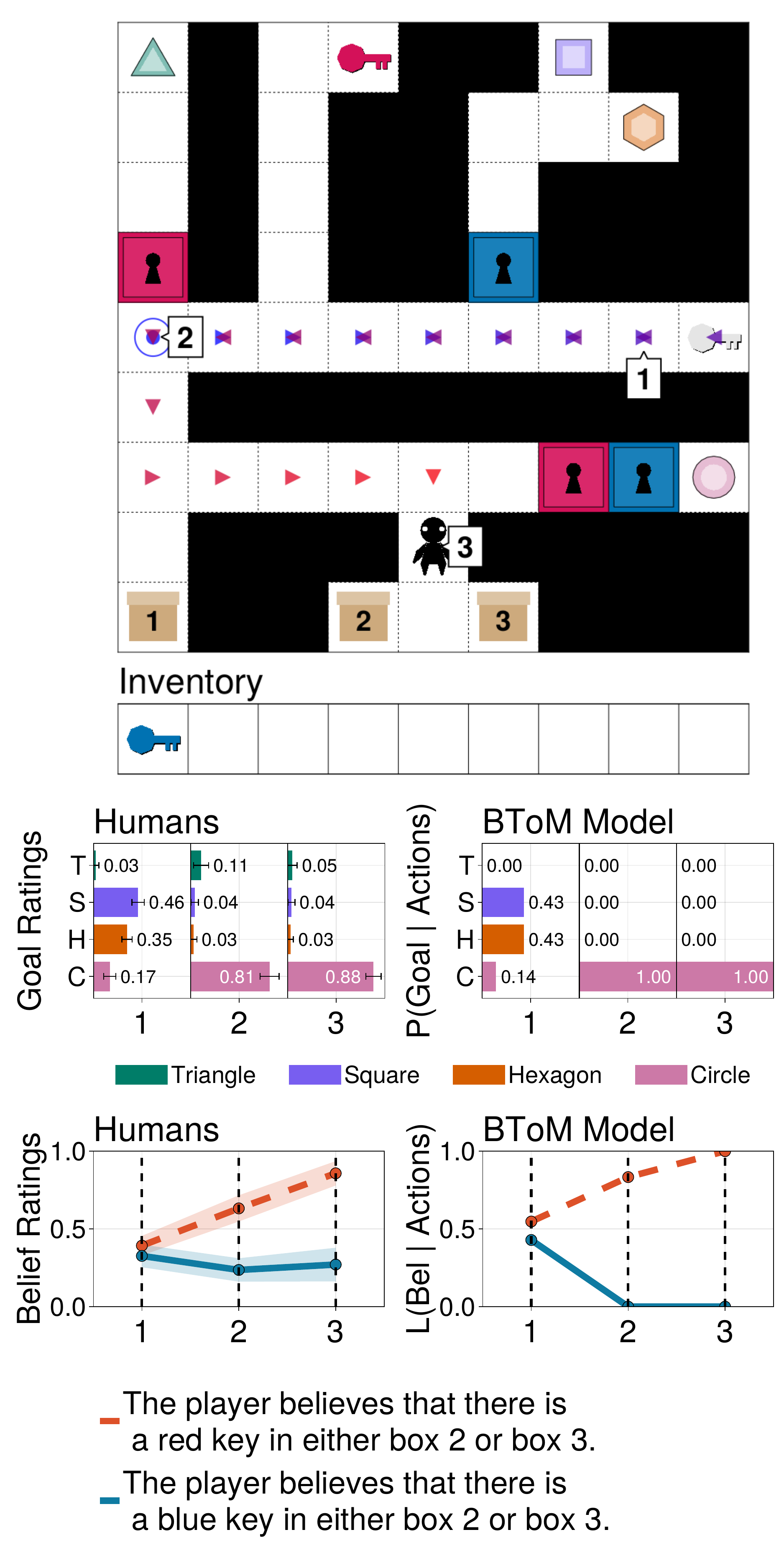}
    \includegraphics[width=0.3\textwidth]{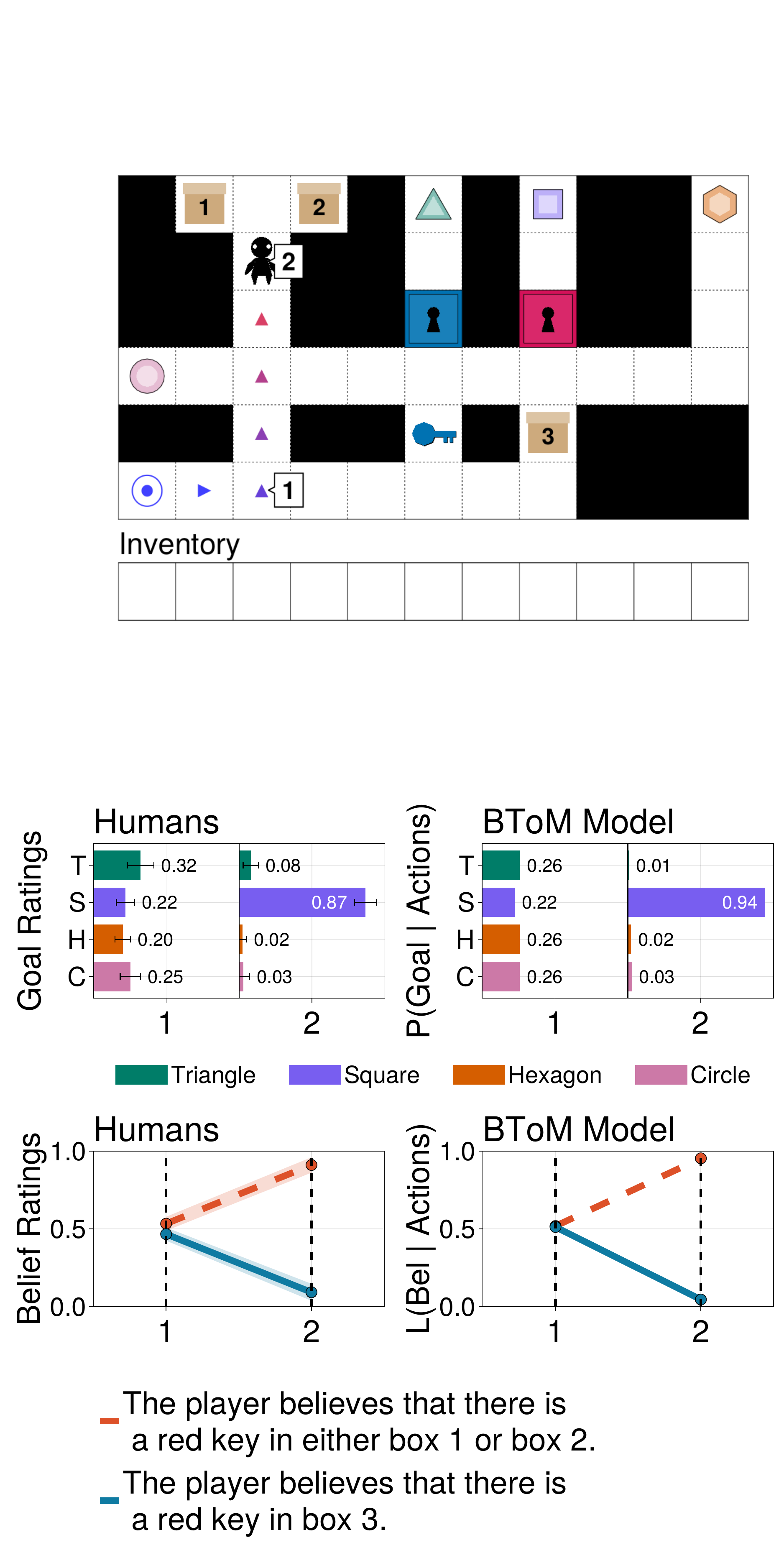}    \includegraphics[width=0.3\textwidth]{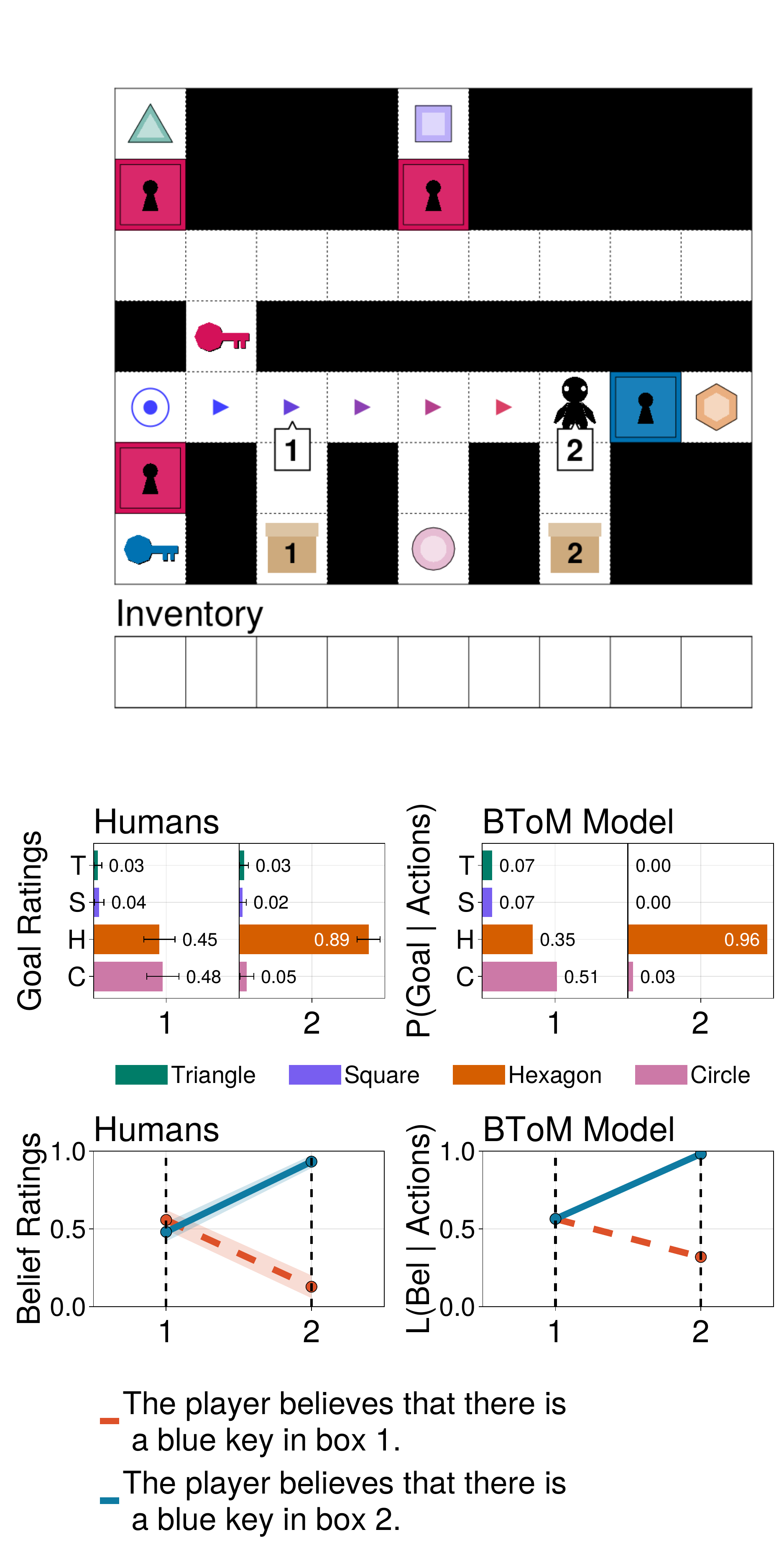}
    \includegraphics[width=0.3\textwidth]{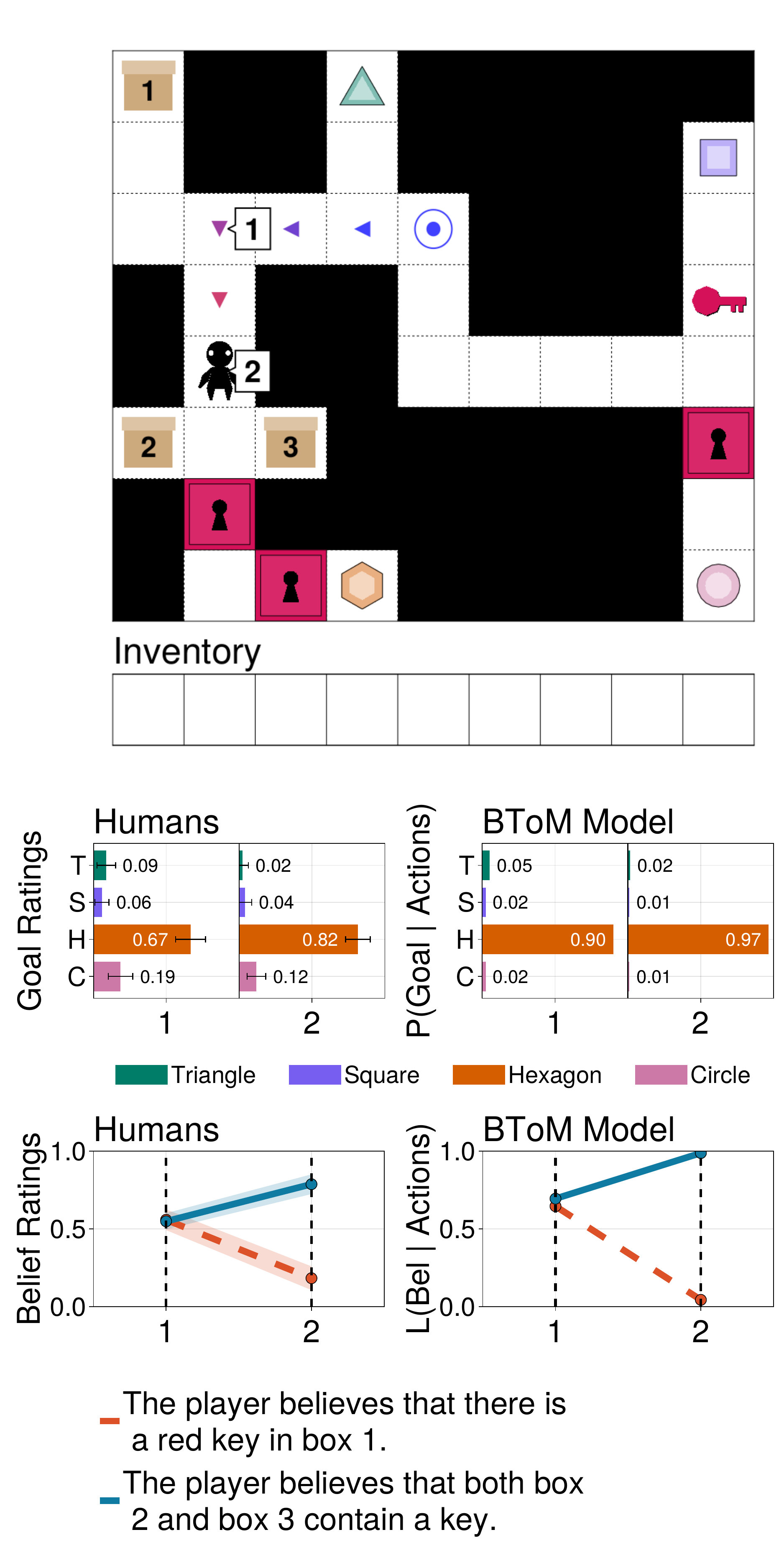}    %
    \caption{Human and model inferences across 6 illustrative scenarios. Judgement points are shown in the map as tooltips annotating the trajectory. Keys picked up along the trajectory are shown in gray. Below each map, we show goal inferences, followed by belief inferences (assuming $U_\psi$ as our belief prior). Overall, our model closely matches human responses.}
    \label{fig:qualitative}
\end{figure*}

\subsection{Scenario Generation}

We constructed 18 scenarios in the Doors, Keys, \& Gems environment with different maze designs and item locations (see Figure \ref{fig:qualitative}). In each scenario, there were 4 goal gems of different shapes (triangle, square, hexagon, circle), and 2 to 3 boxes. Each scenario was paired with two English statements describing the agent's beliefs about the contents of the boxes, such as \emph{``The player believes that there is a red key in box 1.''} To test compositional language understanding, we also created negated, conjunctive, and disjunctive belief statements (e.g. \emph{``The player believes that there is a blue or red key in either box 2 or box 3.''})

\subsection{Experiment Design}

Our study was conducted online through a custom interface. Participants first completed a tutorial and a comprehension quiz. During each scenario, participants rated the goals and beliefs of the observed agent at several judgement points. For goal ratings, participants were presented with a checkbox for each gem, and asked to select all gems that were likely to be the agent's goal. For belief ratings, participants were shown a scale from 1 to 7 below each belief statement, with 1 representing "Definitely False", 7 representing "Definitely True", and 4 representing "Equally Likely to be True or False". We normalized these ratings to lie between 0 and 1.

\subsection{Participants}

We recruited 100 US participants through Prolific (mean age = 39.57, 50 female, 49 male, 1 agender). Each participant rated 9 out of the 18 scenarios. Participants were paid US\$15/hr, and received a bonus for correctly inferring the agents' goals ($\$0.05/n$ for each judgment point where they selected $n$ goals, one of which was correct).


\begin{table*}[ht]
\centering
\caption{Human-model correlations for both goal attributions and belief statement ratings. Bootstrap confidence intervals with a 95\% confidence level are reported in brackets.}
\label{tab: corr}
\begin{tabular}{lccc}
\hline
\multicolumn{1}{c}{\textbf{Model}} & \textbf{Belief Prior} & \textbf{Goal Attributions} & \textbf{Belief Statements} \\
 \hline
Full BToM & Uniform over Statements ($U_\psi$) & \multirow{2}{*}{0.93 [0.91, 0.94]}  & 0.92 [0.91, 0.93] \\
Full BToM & Uniform over States ($U_{S_0}$) &  & 0.86 [0.85, 0.87] \\
Heuristic Mentalizer & Uniform over Statements ($U_\psi$) & \multirow{2}{*}{0.17[0.13, 0.19]} & 0.04 [0.02, 0.06] \\
Heuristic Mentalizer & Uniform over States ($U_{S_0}$) & & 0.19 [-0.03, 0.40] \\
Non-Mentalizer & Uniform over States ($U_{S_0}$) & ---  & 0.19 [-0.03, 0.40]  \\
Omniscient Observer & --- & --- & 0.64 [0.48, 0.75]\\
Ignorant Observer & --- & --- & --- \\
\hline
\end{tabular}
\end{table*}

\begin{figure*}[ht]
    \centering
    \hspace*{-0.5cm}
    \includegraphics[width = 18cm]{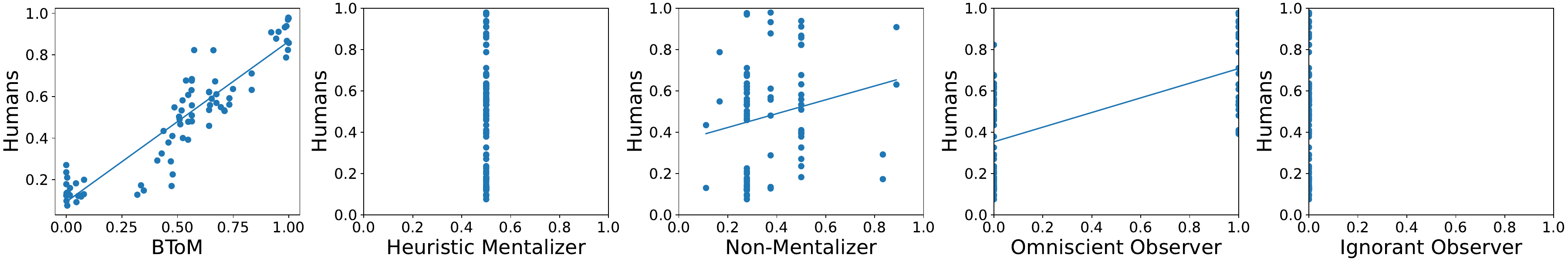}
    \caption{Correlation plots comparing belief judgments from humans ($y$-axis) against models ($x$-axis). Our full BToM model shows a significant better fit to human belief ratings than alternative models}
    \label{fig:corr_plot}
\end{figure*}

\subsection{Alternative Models}

Since our account of the semantics of belief crucially requires the ability to infer the mental states of other agents, we compare our model against baselines that either do not perform such mentalizing at all, or perform limited versions:

\vspace{-10pt}
\paragraph{Omniscient Observer:} Possesses complete knowledge of the validity of the belief statements. This models what epistemic logic would deduce about the agent's beliefs, given the premise that the agent knows everything about the world, and given that the observer does too.

\paragraph{Ignorant Observer:} Rates all belief statements as false due to insufficient premises. This models what epistemic logic would (fail to) conclude about the agent's beliefs, following a semantics of negation as failure to deduce facts. 
\vspace{-10pt}
\paragraph{(Uncertain) Non-Mentalizing Observer:} Rates all belief statements according to the prior probability that they are true, according to the uniform prior $U_{S_0}$ over states.
\vspace{-10pt}
\paragraph{Heuristic Mentalizer:} Assumes that the agents will always move physically closer to their goal, ignoring instrumental actions like picking up keys. This baseline tests the importance of accounting for means-ends coherence when inferring an agent's beliefs from their actions.


\section{Results}

\subsection{Qualitative Analysis}

Figure \ref{fig:qualitative} shows the inferences of both humans and our model in 6 illustrative scenarios. These examples demonstrate qualitatively how our model fits human data. Due to space constraints, we describe only the first 3 examples.

In the first two examples, once the player picks up a red key, both human participants and our model rate the circle to be the only likely goal. This is because the circle is the only goal that requires picking up a red key. Once the player walks down the corridor past box 1, both humans and our model rate statement 1 (\emph{``The player believes that there is a blue key in box 1.''}) to be highly improbable. This is because the player needs a blue key to reach the inferred goal and a rational agent would not walk past box 1 if they knew it contained a blue key. As the agent walks towards box 2 in Example 1, both humans and our model find it likely that there is a blue key in box 2, whereas in Example 2, once the player walks past box 2, both human participants and our model assign a higher likelihood for statement 2 (\emph{``The player believes that there is a blue key in box 3.''})

In Example 3, once the player picks up a blue key, both humans and our model infer the square and the hexagon to be the likely goals, since they are locked behind a blue door nearby. As the player walks past the blue door and towards boxes 2 and 3, these inferences veer towards the circle. It also becomes more likely for there to be a red key than a blue key among boxes 2 and 3, since the player needs a red key to reach the goal. 


\subsection{Quantitative Analysis}

Correlation plots between human and model belief judgments are shown in Figure \ref{fig:corr_plot}, and a summary of key results is presented in Table \ref{tab: corr}. We find that our full BToM model has a much better fit with human ratings than the baselines. This demonstrates the importance of a theory-of-mind for a semantics of belief statements: Without accounting for the coherence between an agent's beliefs, goals, and actions, observers cannot infer an agent's beliefs, and hence cannot evaluate the probability of such belief statements. Our results also demonstrate the importance of a \emph{probabilistic} theory-of-mind. Human judgments about what agents believe are not all-or-nothing phenomena. Our account provides a semantics for such graded judgments, unlike logical models which judge statements to be either false or true.

Interestingly, although both versions of our full model correlated well with human belief judgments (Table \ref{tab: corr}, rows 1--2), the model assuming a uniform statement prior $U_\psi$ had a significantly better fit. We found that this was because certain statements are true in \emph{more possible worlds} (e.g. \emph{"The agent believes that there is a red key in box 1 or box 2."} vs \emph{"The agent believes that there is a red key in box 1."}), such that they have a higher \emph{base rate} of being true if all states $s_0$ are equally probable. Our participants did not seem to take into account these base rates when providing responses. Instead, they tended to rate a statement more highly only if there was more \emph{evidence} for that statement, consistent with our earlier suggestion that belief ratings might correspond to a normalized likelihood $\bar L(\psi | a_{1:T})$. While this is not Bayesian in the orthodox sense, it is semantically and pragmatically quite reasonable --- it suggests that people are only willing to say that an agent believes $\phi$ if there is evidence for that belief.

\section{Discussion}

In this study, we explored how humans reason about other agents' beliefs and interpret beliefs communicated in natural language. Our experiment shows the deep connection be-tween how we interpret language about belief and how we understand other agents' minds: People are able to interpret and adjust their evaluations of natural language statements as new actions come to light, demonstrating the importance of grounding such language in a theory of how agents' beliefs and goals are connected to their plans and actions. Such a theory provides a functional role for belief as a concept.

All that said, there is much work to be done before a model like ours can account for all the ways in which people use and interpret language about beliefs.
A key limitation of our current model is that it only explains the \emph{deterministic} beliefs of other agents, under the assumption that what other agents believe is what they \emph{know}. But of course, one of the hallmarks about belief --- and how we reason and talk about it --- is that it can come apart from reality in all sorts of ways. People have uncertain beliefs (just like our observer), which we express in language using modals like "might" or "unlikely".  People have false beliefs, and we form sentences describing them all the time (\emph{``He thinks that ..., but actually...''}). People might be ignorant or possess only partial knowledge, and we describe this differently from mere uncertainty. Relatedly, people often think and talk about belief in abstract terms, without concretizing those beliefs into concrete worlds like our model does. If we are to explain such phenomena, we will need a much richer theory-of-mind than our model currently offers: One that models our uncertainty about other's uncertainty \shortcite{baker2017rational,gmytrasiewicz2005framework}, as well as the abstract ways we represent both the world and each others' minds \shortcite{bigelow2023non}.


\section{Acknowledgements}
This work was funded in part by the DARPA Machine Common
Sense, AFOSR, and ONR Science of AI programs, along with the
MIT-IBM Watson AI Lab, and gifts from Reid Hoffman and the
Siegel Family Foundation. Tan Zhi-Xuan is supported by an Open
Philanthropy AI Fellowship.

We thank our colleagues Brian Leahy, Cedegao Zhang, and Megan Wei for helpful discussions and suggestions during the development of this project.

\bibliographystyle{apacite}

\setlength{\bibleftmargin}{.125in}
\setlength{\bibindent}{-\bibleftmargin}

\bibliography{paper}

\end{document}